\documentclass{article}

\usepackage{arxiv}
\usepackage[utf8]{inputenc}
\usepackage[T1]{fontenc}
\usepackage[colorlinks=true, linkcolor=snsblue, citecolor=snsgreen, urlcolor=snsred]{hyperref}
\usepackage{url}
\usepackage{booktabs}
\usepackage{amsmath}
\usepackage{amssymb}
\usepackage{amsfonts}
\usepackage{nicefrac}
\usepackage{microtype}
\usepackage{natbib}
\usepackage{siunitx}
\usepackage{mathtools}
\usepackage{algorithm}
\usepackage{algpseudocode}
\usepackage[nameinlink]{cleveref}

\usepackage{xcolor}
\definecolor{snsblue}{RGB}{76, 114, 176}
\definecolor{snsgreen}{RGB}{85, 168, 104}
\definecolor{snsred}{RGB}{196, 78, 82}
\definecolor{snspurple}{RGB}{129, 114, 178}
\definecolor{snsyellow}{RGB}{204, 185, 116}
\definecolor{snscyan}{RGB}{100, 181, 205}

\newcommand{\labelSC}{\texttt{SC}}
\newcommand{\labelWC}{\texttt{WC}}
\newcommand{\labelSCNNt}{\texttt{SC-NNt}}
\newcommand{\labelSCNNa}{\texttt{SC-NNa}}
\newcommand{\labelNN}{\texttt{NN}}
\newcommand{\ie}{i.e.}
\newcommand{\eg}{e.g.}

\title{Online model error correction with neural networks in the incremental 4D-Var framework}

\author{
Alban Farchi\\
CEREA, École des Ponts and EDF R\&D\\
\^Ile--de--France, France\\
\texttt{alban.farchi@enpc.fr}\\
\AND
Marcin Chrust\\
ECMWF\\
Shinfield Park\\
Reading, United Kingdom\\
\And
Marc Bocquet\\
CEREA, École des Ponts and EDF R\&D\\
\^Ile--de--France, France\\
\AND
Patrick Laloyaux\\
ECMWF\\
Shinfield Park\\
Reading, United Kingdom\\
\And
Massimo Bonavita\\
ECMWF\\
Shinfield Park\\
Reading, United Kingdom
}

\begin{document}

\maketitle

\begin{abstract}
Recent studies have demonstrated that it is possible to combine machine learning with data assimilation to reconstruct the dynamics of a physical model partially and imperfectly observed. Data assimilation is used to estimate the system state from the observations, while machine learning computes a surrogate model of the dynamical system based on those estimated states. The surrogate model can be defined as an hybrid combination where a physical model based on prior knowledge is enhanced with a statistical model estimated by a neural network. The training of the neural network is typically done offline, once a large enough dataset of model state estimates is available. By contrast, with online approaches the surrogate model is improved each time a new system state estimate is computed. Online approaches naturally fit the sequential framework encountered in geosciences where new observations become available with time. In a recent methodology paper, we have developed a new weak-constraint 4D-Var formulation which can be used to train a neural network for online model error correction. In the present article, we develop a simplified version of that method, in the incremental 4D-Var framework adopted by most operational weather centres. The simplified method is implemented in the ECMWF Object-Oriented Prediction System, with the help of a newly developed Fortran neural network library, and tested with a two-layer two-dimensional quasi geostrophic model. The results confirm that online learning is effective and yields a more accurate model error correction than offline learning. Finally, the simplified method is compatible with future applications to state-of-the-art models such as the ECMWF Integrated Forecasting System.
\end{abstract}


\section*{Plain language summary}

We have recently proposed a general framework for combining data assimilation and machine learning techniques to train a neural network for online model error correction. In the present article, we develop a simplified version of this online training method, compatible with future applications to more realistic models. Using numerical illustrations, we show that the new method is effective and yields a more accurate model error correction than the usual offline learning approach. The results show the potential of incorporating data assimilation and machine learning tightly, and pave the way towards an application to the Integrated Forecasting System used for operational numerical weather prediction at the European Centre for Medium-Range Weather Forecasts.

\section*{Key points}

\begin{itemize}
    \item Weak-constraint 4D-Var variants can be used to train neural networks for online model error correction.
    \item Online learning yields a more accurate model error correction than offline learning.
    \item The new, simplified method, developed in the incremental 4D-Var framework, can be easily applied in operational weather models.
\end{itemize}

\section{Introduction: machine learning for model error correction}

In the geosciences, data assimilation (DA) is used to increase the quality of forecasts by providing accurate initial conditions \citep{kalnay2003, reich-2015, law-2015, asch-2016, carrassi-2018, evensen2022}. The initial conditions are obtained by combining all sources of information in a mathematically optimal way, in particular information from the dynamical model and information from sparse and noisy observations. There are two main classes of DA methods. In variational DA, the core of the methods is to minimise a cost function, usually using gradient-based optimisation techniques, to estimate the system state. Examples include 3D- and 4D-Var. In statistical DA, the methods relies on the sampled error statistics to perform sequential updates to the state estimation. The most popular examples are the ensemble Kalman filter (EnKF) and the particle filter.

Most of the time, DA methods are applied with the perfect model assumption: this is called strong-constraint DA. However, despite the significant effort provided by the modellers, geoscientific models remain affected by errors \citep{dee-2005}, for example due to unresolved small-scale processes. This is why there is a growing interest of the DA community in weak-constraint (WC) methods, \ie{} DA methods relaxing the perfect model assumption \citep{tremolet-2006}. This has led, for example, to the iterative ensemble Kalman filter in the presence of additive noise \citep{sakov-2018} in statistical DA, and to the forcing formulation of WC 4D-Var \citep{laloyaux-2020} in variational DA. In practice, the DA control vector has to be extended to include the model error in addition to the system state. The downside of this approach is the potentially significant increase of the problem's dimension since the model trajectory is not anymore described uniquely by the initial condition. By construction, WC 4D-Var is an online model error correction method, meaning that the model error is estimated during the assimilation process and only valid for the states in the current assimilation window.

In parallel, following the renewed impetus of machine learning (ML) applications \citep{lecun-2015, goodfellow-2016, chollet-2018}, data-driven approaches are more and more frequent in the geosciences. The goal of these approaches \citep[\textit{e.g.,}][among many others]{brunton-2016, hamilton-2016, lguensat-2017, pathak-2018, dueben-2018, fablet-2018, scher-2019, weyn-2019, arcomano-2020} is to learn a surrogate of the dynamical model using supervised learning, \ie{} by minimising a loss function which measures the discrepancy between the surrogate model predictions and an observation dataset. In order to take into account sparse and noisy observations, ML techniques can be combined with DA \citep{abarbanel-2018, bocquet-2019a, brajard-2020, bocquet-2020, arcucci-2021}. The idea is to take the best of both worlds: DA techniques are used to estimate the state of the system from the observations, and ML techniques are used to estimate the surrogate model from the estimated state. In practice, the hybrid DA and ML methods can be used both for full model emulation and model error correction \citep{rasp-2018, pathak-2018a, bolton-2019, jia-2019, watson-2019, bonavita-2020, brajard-2020b, gagne-2020, wikner-2020, farchi-2021b, farchi-2021, chen-2022}. In the first case, the surrogate model is entirely learned from observations, while in the latter case, the surrogate model is hybrid: a physical, knowledge-based model is corrected by a statistical model, \eg{} a neural network (NN), which is learned from observations. Even though from a technical point of view it can arguably be more difficult to implement, model error correction has many advantages over full model emulation: by leveraging the long history of numerical modelling, one can hope to end up with an easier learning problem \citep{watson-2019, farchi-2021}.

Most of the current hybrid DA-ML methods use offline learning strategies: the surrogate model (or model error correction) is learned using a large dataset of observations (or analyses) and should be generalisable to other situations (\ie{} outside the dataset). There are two main reasons for this choice. First, surrogate modelling requires a large amount of data to provide accurate results -- certainly more than what is available in a single assimilation update with online learning. Second, by doing so, it is possible to use the full potential of the ML variational tools. Nevertheless, online learning has on paper several advantages over offline learning. 
\begin{itemize}
    \item Online learning fits the standard sequential DA approach in the geosciences. Each time a new batch of observations becomes available, the surrogate model parameters can be corrected.
    \item With online learning, the system state and the surrogate model parameters are jointly estimated, which is often not the case with offline learning. Joint estimation is in general more consistent, and hence potentially more accurate, than separate estimation.
    \item With offline learning, the training only starts once a sufficiently large dataset is available. With online learning, the training begins from the first batch of observations, which means that improvements can be expected before having a large dataset.
    \item With online learning, since the surrogate model is constantly updated, it can adapt to new (previously unseen) conditions. An example could be, in the case of model error correction, an update of the physical model to correct. Another example could be slowly-varying effects on the dynamics (\eg, seasonality).
\end{itemize}
Fundamentally, online learning is very similar to parameter estimation in DA: the goal is to estimate at the same time the system state and some parameters -- in this case the surrogate model parameters. Several example of online learning methods have recently emerged. \citet{bocquet-2020a, malartic-2022} have developed several variants of the EnKF to perform a joint estimation of the state and the parameters of surrogate model which fully emulates the dynamics. \citet{gottwald-2021} have used a very similar approach for the parameters of an echo state network used as surrogate model. Finally, \citet{farchi-2021b} have derived a variant of WC 4D-Var to perform a joint estimation of the state and the parameters of a NN which correct the tendencies of a physical model. 
In this article, we revisit the method of \citet{farchi-2021b}. A new simplified method is derived, compatible with future applications to more realistic models. The method is implemented in the Object-Oriented Prediction System (OOPS) framework developed at the European Center for Medium-Range Weather Forecasts (ECMWF), and tested using the two-layer quasi-geostrophic channel model developed in OOPS. To us, this is a final step before considering an application with the Integrated Forecasting System \citep[IFS,][]{bonavita-2017}, since the IFS will soon rely on OOPS for its DA part.

The article is organised as follows. \Cref{sec:methodology} presents the methodology. The quasi-geostrophic (QG) model is described in \cref{sec:qg}. The experimental results are then presented in \cref{sec:offline} for offline learning, and in \cref{sec:online} for online learning. Finally, conclusions are given in \cref{sec:conclusions}.

\section{A simplified neural network variant of weak-constraint 4D-Var}
\label{sec:methodology}

\subsection{Strong-constraint 4D-Var}
\label{ssec:methodology-sc4dvar}

Suppose that we follow the evolution of a system using a series of observations taken at discrete times. In the classical 4D-Var, the observations are gathered into time windows $\left(\mathbf{y}_{0}, \ldots, \mathbf{y}_{L}\right)$. The integer $L\ge0$ is the window length, and $\mathbf{y}_{k}\in\mathbb{R}^{N_{\mathsf{y}}}$, the $k$-th batch of observations, contains all the observations taken at time $t_{k}$, for $k=0, \ldots, L$. For convenience, we assume that the time interval between consecutive observation batches $t_{k+1}-t_{k}=\Delta t$ is constant. This assumption is not fundamental; it just makes the presentation much easier. Within the window, the system state $\mathbf{x}_{k}\in\mathbb{R}^{N_{\mathsf{x}}}$ at time $t_{k}$ is obtained by integrating the model in time from $t_{0}$ to $t_{k}$:
\begin{equation}
    \label{eq:methodology-sc4dvar-model}
    \mathbf{x}_{k} = \boldsymbol{\mathcal{M}}_{k:0}\left(\mathbf{x}_{0}\right),
\end{equation}
where $\boldsymbol{\mathcal{M}}_{k:l}:\mathbb{R}^{N_{\mathsf{x}}}\to\mathbb{R}^{N_{\mathsf{x}}}$ is the resolvent of the dynamical (or physical) model from $t_{l}$ to $t_{k}$. Moreover, the observations are related to the state by the observation operator $\boldsymbol{\mathcal{H}}_{k}:\mathbb{R}^{N_{\mathsf{x}}}\to\mathbb{R}^{N_{\mathsf{y}}}$ via
\begin{equation}
    \mathbf{y}_{k} = \boldsymbol{\mathcal{H}}_{k}\left(\mathbf{x}_{k}\right) + \mathbf{v}_{k},
\end{equation}
where $\mathbf{v}_{k}$ is the observation error at time $t_{k}$, which could be a random vector. Let us make the assumption that the observation errors are independent from each other.

The 4D-Var cost function is defined as the negative log-likelihood:
\begin{subequations}
    \begin{align}
        \mathcal{J}^{\mathsf{sc}}\left(\mathbf{x}_{0}\right) &\triangleq -\ln p\left(\mathbf{x}_{0}|\mathbf{y}_{0}, \ldots, \mathbf{y}_{L}\right),\\
        &\propto -\ln p\left(\mathbf{x}_{0}\right) -\ln p\left(\mathbf{y}_{0}, \ldots, \mathbf{y}_{L}|\mathbf{x}_{0}\right),\\
        &\propto -\ln p\left(\mathbf{x}_{0}\right) -\sum_{k=0}^{L}\ln p\left(\mathbf{y}_{k}|\mathbf{x}_{0}\right),
    \end{align}
\end{subequations}
where conditional independence of the observation vectors on $\mathbf{x}_{0}$ was used.
The background $p\left(\mathbf{x}_{0}\right)$ is Gaussian with mean $\mathbf{x}^{\mathsf{b}}_{0}$ and covariance matrix $\mathbf{B}$, and the observation errors $\mathbf{v}_{k}$ are also Gaussian distributed with mean $\mathbf{0}$ and covariance matrices $\mathbf{R}_{k}$, in such a way that $\mathcal{J}^{\mathsf{sc}}$ becomes:
\begin{equation}
    \label{eq:methodology-sc4dvar-cost-gaussian}
    \mathcal{J}^{\mathsf{sc}}\left(\mathbf{x}_{0}\right) = \frac{1}{2} \left\|\mathbf{x}_{0}-\mathbf{x}^{\mathsf{b}}_{0}\right\|^{2}_{\mathbf{B}^{-1}} + \frac{1}{2} \sum_{k=0}^{L} \left\|\mathbf{y}_{k}-\boldsymbol{\mathcal{H}}_{k}\circ\boldsymbol{\mathcal{M}}_{k:0}\left(\mathbf{x}_{0}\right)\right\|^{2}_{\mathbf{R}^{-1}_{k}},
\end{equation}
where we have dropped the constant terms and where the notation $\left\|\mathbf{v}\right\|^{2}_{\mathbf{M}}$ stands for the squared Mahalanobis norm $\mathbf{v}^{\top}\mathbf{M}\mathbf{v}$.

This formulation is called \emph{strong-constraint} 4D-Var because it relies on the perfect model assumption \cref{eq:methodology-sc4dvar-model}. In practice, \cref{eq:methodology-sc4dvar-cost-gaussian} is minimised using scalable gradient-based optimisation methods to provide the analysis $\mathbf{x}^{\mathsf{a}}_{0}$. In cycled DA, the model is then used to propagate $\mathbf{x}^{\mathsf{a}}_{0}$ till the start of the next window, yielding thus a value for the background state $\mathbf{x}^{\mathsf{b}}_{0}$.

\subsection{Weak-constraint 4D-Var}
\label{ssec:methodology-wc4dvar}

Recognising that the model is not perfect, we can replace the strong constraint \cref{eq:methodology-sc4dvar-model} by the more general model evolution
\begin{equation}
    \label{eq:methodology-wc4dvar-model}
    \mathbf{x}_{k+1} = \boldsymbol{\mathcal{M}}_{k+1:k}\left(\mathbf{x}_{k}\right) + \mathbf{w}_{k},
\end{equation}
where $\mathbf{w}_{k}\in\mathbb{R}^{N_{\mathsf{x}}}$ is the model error from $t_{k}$ to $t_{k+1}$, potentially random. Let us make the assumption that the model errors are independent from each other and independent from the background errors. This implies that the model evolution satisfies the Markov property.

The updated cost function now depends on all states inside the window:
\begin{subequations}
    \begin{align}
        \mathcal{J}^{\mathsf{wc}}\left(\mathbf{x}_{0}, \ldots, \mathbf{x}_{L}\right) &\triangleq -\ln p\left(\mathbf{x}_{0}, \ldots, \mathbf{x}_{L}|\mathbf{y}_{0}, \ldots, \mathbf{y}_{L}\right), \\
        &\propto -\ln p\left(\mathbf{x}_{0}, \ldots, \mathbf{x}_{L}\right) -\ln p\left(\mathbf{y}_{0}, \ldots, \mathbf{y}_{L}|\mathbf{x}_{0}, \ldots, \mathbf{x}_{L}\right),\\
        &\propto -\ln p\left(\mathbf{x}_{0}\right) -\sum_{k=0}^{L-1}\ln p\left(\mathbf{x}_{k+1}|\mathbf{x}_{k}\right) -\sum_{k=0}^{L}\ln p\left(\mathbf{y}_{k}|\mathbf{x}_{k}\right).
    \end{align}
\end{subequations}
With the Gaussian assumptions of \cref{ssec:methodology-sc4dvar} and the additional hypothesis that the model errors $\mathbf{w}_{k}$ also follow a Gaussian distribution with mean $\mathbf{w}^{\mathsf{b}}_{k}$ and covariance matrices $\mathbf{Q}_{k}$, $\mathcal{J}^{\mathsf{wc}}$ becomes
\begin{align}
    \label{eq:methodology-wc4dvar-cost-gaussian}
    \mathcal{J}^{\mathsf{wc}}\left(\mathbf{x}_{0}, \ldots, \mathbf{x}_{L}\right) ={} & \frac{1}{2} \left\|\mathbf{x}_{0}-\mathbf{x}^{\mathsf{b}}_{0}\right\|^{2}_{\mathbf{B}^{-1}} + \frac{1}{2} \sum_{k=0}^{L-1} \left\|\mathbf{x}_{k+1}-\boldsymbol{\mathcal{M}}_{k+1:k}\left(\mathbf{x}_{k}\right)-\mathbf{w}^{\mathsf{b}}_{k}\right\|^{2}_{\mathbf{Q}^{-1}_{k}} \nonumber \\ & + \frac{1}{2} \sum_{k=0}^{L} \left\|\mathbf{y}_{k}-\boldsymbol{\mathcal{H}}_{k}\left(\mathbf{x}_{k}\right)\right\|^{2}_{\mathbf{R}^{-1}_{k}},
\end{align}
where we have once again dropped the constant terms. This formulation is called \emph{weak-constraint} 4D-Var \citep{tremolet-2006} because it relaxes the perfect model assumption \cref{eq:methodology-sc4dvar-model}, which means that the analysis $\left(\mathbf{x}^{\mathsf{a}}_{0}, \ldots, \mathbf{x}^{\mathsf{a}}_{L-1}\right)$ is not any more a trajectory of the model. However, this comes at a price: the dimension of the problem has increased from $N_{\mathsf{x}}$ to $LN_{\mathsf{x}}$. 

This dimensionality increase can be mitigated by making additional assumptions. For example, one can assume that the model error is constant throughout the window, \ie
\begin{subequations}
    \begin{align}
        \mathbf{w}_{0}=\ldots=\mathbf{w}_{L-1}&\triangleq\mathbf{w},\\
        \mathbf{w}^{\mathsf{b}}_{0}=\ldots=\mathbf{w}^{\mathsf{b}}_{L-1}&\triangleq\mathbf{w}^{\mathsf{b}},\\
        \mathbf{Q}_{0}=\ldots=\mathbf{Q}_{L-1}&\triangleq L\mathbf{Q}.
    \end{align}
\end{subequations}
In this case, the trajectory $\left(\mathbf{x}_{0}, \ldots, \mathbf{x}_{L}\right)$ is fully determined by $\left(\mathbf{w}, \mathbf{x}_{0}\right)$:
\begin{align}
    \label{eq:methodology-wc4dvar-model-forcing}
    \mathbf{x}_{k} = \boldsymbol{\mathcal{M}}_{k+1:k}\left(\mathbf{x}_{k}\right)+\mathbf{w} = \boldsymbol{\mathcal{M}}_{k+1:k}\left(
    \boldsymbol{\mathcal{M}}_{k:k-1}\left(\mathbf{x}_{k-1}\right)+\mathbf{w}
    \right)+\mathbf{w} = \ldots \triangleq \boldsymbol{\mathcal{M}}^{\mathsf{wc}}_{k+1:0}\left(\mathbf{w}, \mathbf{x}_{0}\right),
\end{align}
with $\mathbf{x}\mapsto\boldsymbol{\mathcal{M}}^{\mathsf{wc}}_{k+1:0}\left(\mathbf{w}, \mathbf{x}\right)$ being the \emph{resolvent} of the $\mathbf{w}$-\emph{debiased} model from $t_{0}$ to $t_{k+1}$. The Gaussian cost function $\mathcal{J}^{\mathsf{wc}}$ \cref{eq:methodology-wc4dvar-cost-gaussian} can hence be written
\begin{equation}
    \label{eq:methodology-wc4dvar-forcing-cost-gaussian}
    \mathcal{J}^{\mathsf{wc}}\left(\mathbf{w}, \mathbf{x}_{0}\right) = \frac{1}{2} \left\|\mathbf{x}_{0}-\mathbf{x}^{\mathsf{b}}_{0}\right\|^{2}_{\mathbf{B}^{-1}} + \frac{1}{2}\left\|\mathbf{w}-\mathbf{w}^{\mathsf{b}}\right\|^{2}_{\mathbf{Q}^{-1}} + \frac{1}{2} \sum_{k=0}^{L} \left\|\mathbf{y}_{k}-\boldsymbol{\mathcal{H}}_{k}\circ\boldsymbol{\mathcal{M}}^{\text{wc}}_{k:0}\left(\mathbf{w}, \mathbf{x}_{0}\right)\right\|^{2}_{\mathbf{R}^{-1}_{k}}.
\end{equation}
This approach is called \emph{forcing} formulation of WC 4D-Var \citep{tremolet-2006, fisher-2011, laloyaux-2020} and is the one that is implemented at ECMWF \citep{laloyaux-2020b}. By construction, the perfect model assumption \cref{eq:methodology-sc4dvar-model} is relaxed, but the analysis $\left(\mathbf{w}^{\mathsf{a}}, \mathbf{x}^{\mathsf{a}}_{0}\right)$
yields a trajectory of the $\mathbf{w}^{\mathsf{a}}$-debiased model. In cycled DA, this $\mathbf{w}^{\mathsf{a}}$-debiased model is used to propagate $\mathbf{x}^{\mathsf{a}}_{0}$ until the start of the next window to provide the background state $\mathbf{x}^{\mathsf{b}}_{0}$. However this time, a background is also needed for model error $\mathbf{w}^{\mathsf{b}}$. The simplest option is to use $\mathbf{w}^{\mathsf{a}}$ as is, in other words make the assumption that the dynamical model for model error is persistence.

Hereafter, the forcing formulation of WC 4D-Var is simply called WC 4D-Var.

\subsection{A neural network formulation of weak-constraint 4D-Var}
\label{ssec:methodology-nn4dvar}

Following the approach of \citet{farchi-2021b}, we now assume that the dynamical model is parametrised by a set of parameters $\mathbf{p}\in\mathbb{R}^{N_{\mathsf{p}}}$ constant over the window, in such a way that the model integration \cref{eq:methodology-sc4dvar-model} becomes
\begin{equation}
    \label{eq:methodology-nn4dvar-model}
    \mathbf{x}_{k} = \boldsymbol{\mathcal{M}}^{\mathsf{nn}}_{k:0}\left(\mathbf{p}, \mathbf{x}_{0}\right),
\end{equation}
where $\mathbf{x}\mapsto\boldsymbol{\mathcal{M}}^{\mathsf{nn}}_{k:0}\left(\mathbf{p}, \mathbf{x}\right)$ is the resolvent of the $\mathbf{p}$-\emph{parametrised} model from $t_{0}$ to $t_{k}$. Using the state augmentation principle \citep{jazwinski-1970}, the model parameters $\mathbf{p}$ can be included in the control variables and hence be estimated in DA. If we further assume that the background for model parameters and system state are independent, and that the background for model parameters is Gaussian with mean $\mathbf{p}^{\mathsf{b}}$ and covariance matrix $\mathbf{P}$, then the Gaussian cost function \cref{eq:methodology-sc4dvar-cost-gaussian} becomes
\begin{equation}
    \label{eq:methodology-nn4dvar-cost-gaussian}
    \mathcal{J}^{\mathsf{nn}}\left(\mathbf{p}, \mathbf{x}_{0}\right) = \frac{1}{2} \left\|\mathbf{x}_{0}-\mathbf{x}^{\mathsf{b}}_{0}\right\|^{2}_{\mathbf{B}^{-1}} + \frac{1}{2} \left\|\mathbf{p}-\mathbf{p}^{\mathsf{b}}\right\|^{2}_{\mathbf{P}^{-1}} + \frac{1}{2} \sum_{k=0}^{L} \left\|\mathbf{y}_{k}-\boldsymbol{\mathcal{H}}_{k}\circ\boldsymbol{\mathcal{M}}^{\mathsf{nn}}_{k:0}\left(\mathbf{p}, \mathbf{x}_{0}\right)\right\|^{2}_{\mathbf{R}^{-1}_{k}}.
\end{equation}
This formulation is called \emph{neural network} 4D-Var because in the present article, the set of parameters $\mathbf{p}$ are typically the weights and biases of a NN. Nevertheless, we would like to emphasise the fact that this formulation is not restricted only to NNs and can be used to estimate any parameters. The similarity between \cref{eq:methodology-wc4dvar-forcing-cost-gaussian,eq:methodology-nn4dvar-cost-gaussian} is clear, which is why NN 4D-Var should be seen as another formulation of WC 4D-Var. By construction, the perfect model assumption \cref{eq:methodology-sc4dvar-model} is once again relaxed, but this time the analysis $\left(\mathbf{p}^{\mathsf{a}}, \mathbf{x}^{\mathsf{a}}_{0}\right)$ yields a trajectory of the $\mathbf{p}^{\mathsf{a}}$-parametrised model. In cycled DA, this $\mathbf{p}^{\mathsf{a}}$-parametrised model is used to propagate the analysis state $\mathbf{x}^{\mathsf{a}}_{0}$ until the start of the next window to provide the background state $\mathbf{x}^{\mathsf{b}}_{0}$. Once again, a background is also needed for model parameters $\mathbf{p}^{\mathsf{b}}$. The simplest option is to use $\mathbf{p}^{\mathsf{a}}$ as is, in other words make the assumption that the evolution model for model parameters is persistence.

Even though there are a lot of similarities between NN 4D-Var and the WC 4D-Var, two essential differences should be highlighted:
\begin{enumerate}
    \item The model error $\mathbf{w}$ lies in the state space $\mathbb{R}^{N_{\mathsf{x}}}$ while the model parameters lies in the parameter space $\mathbb{R}^{N_{\mathsf{p}}}$, which has consequences on the design of the covariance matrices $\mathbf{Q}\in\mathbb{R}^{N_{\mathsf{x}}\times N_{\mathsf{x}}}$ and $\mathbf{P}\in\mathbb{R}^{N_{\mathsf{p}}\times N_{\mathsf{p}}}$.
    \item More importantly, $\boldsymbol{\mathcal{M}}^{\mathsf{wc}}_{k:0}$ and $\boldsymbol{\mathcal{M}}^{\mathsf{nn}}_{k:0}$ may have different functional forms. In particular, in the first case the model error $\mathbf{w}$ is constant while in the second case, it is the model parameters $\mathbf{p}$ which are constant.
\end{enumerate}

\subsection{A simplified NN 4D-Var for model error correction}
\label{ssec:methodology-nn4dvar-simplified}

In the present article, we want to use NN 4D-Var for model error correction. Let us consider the case where the parametrised model is written
\begin{equation}
    \label{eq:methodology-nn4dvar-simplified-model}
    \mathbf{x}_{k+1} = \boldsymbol{\mathcal{M}}^{\mathsf{nn}}_{k+1:k}\left(\mathbf{p}, \mathbf{x}_{k}\right) = \boldsymbol{\mathcal{M}}_{k+1:k}\left(\mathbf{x}_{k}\right) + \boldsymbol{\mathcal{F}}\left(\mathbf{p}, \mathbf{x}_{k}\right),
\end{equation}
where $\boldsymbol{\mathcal{F}}$ is a NN correction added to $\boldsymbol{\mathcal{M}}_{k+1:k}$, the resolvent of the (non-corrected) physical model from $t_{k}$ to $t_{k+1}$, and $\mathbf{p}$ are the parameters of this NN. Following the approach of \cref{ssec:methodology-wc4dvar}, we assume that the NN is autonomous, \ie{} the NN correction is constant throughout the window. The model evolution \cref{eq:methodology-nn4dvar-simplified-model} can hence be written 
\begin{equation}
    \boldsymbol{\mathcal{M}}^{\mathsf{nn}}_{k+1:k}\left(\mathbf{p}, \mathbf{x}_{k}\right) = \boldsymbol{\mathcal{M}}_{k+1:k}\left(\mathbf{x}_{k}\right) + \mathbf{w}, \quad \mathbf{w}=\boldsymbol{\mathcal{F}}\left(\mathbf{p}, \mathbf{x}_{0}\right).
\end{equation}
This evolution model can then be plugged into the cost function $\mathcal{J}^{\mathsf{nn}}$ \cref{eq:methodology-nn4dvar-cost-gaussian}, which yields a simplified variant of NN 4D-Var where the NN is used only once per cycle. Furthermore, comparing this to \cref{eq:methodology-wc4dvar-model-forcing}, we conclude that
\begin{equation}
    \boldsymbol{\mathcal{M}}^{\mathsf{nn}}_{k:0}\left(\mathbf{p}, \mathbf{x}_{0}\right) = \boldsymbol{\mathcal{M}}^{\mathsf{wc}}_{k:0}\left(\boldsymbol{\mathcal{F}}\left(\mathbf{p}, \mathbf{x}_{0}\right), \mathbf{x}_{0}\right).
\end{equation}
This means that it will be possible to build this new method on top of the currently implemented WC 4D-Var framework, which is a major practical advantage.

In practice, the minimisation method implemented at ECMWF relies on an incremental approach with \emph{outer} and \emph{inner} loops \citep{courtier-1994}. In each outer loop, the cost function is linearised about the first-guess, and the linearised cost function is then minimised in the inner loop, typically using the conjugate gradient algorithm. Let us see how this works for our simplified NN 4D-Var. Using the change of variables $\left(\delta\mathbf{p}, \delta\mathbf{x}_{0}\right)\triangleq\left(\mathbf{p}-\mathbf{p}^{\mathsf{i}}, \mathbf{x}_{0}-\mathbf{x}^{\mathsf{i}}_{0}\right)$, where $\left(\mathbf{p}^{\mathsf{i}}, \mathbf{x}^{\mathsf{i}}_{0}\right)$ is the first guess, we have
\begin{subequations}
    \begin{align}
        \mathcal{J}^{\mathsf{nn}}\left(\mathbf{p}, \mathbf{x}_{0}\right) ={} & {\mathcal{J}^{\mathsf{nn}}\left(\mathbf{p}^{\mathsf{i}}+\delta\mathbf{p}, \mathbf{x}^{\mathsf{i}}_{0}+\delta\mathbf{x}_{0}\right),}\\
        ={} &{\frac{1}{2} \left\|\mathbf{x}^{\mathsf{i}}_{0}-\mathbf{x}^{\mathsf{b}}_{0}+\delta\mathbf{x}_{0}\right\|^{2}_{\mathbf{B}^{-1}} + \frac{1}{2} \left\|\mathbf{p}^{\mathsf{i}}-\mathbf{p}^{\mathsf{b}}+\delta\mathbf{p}\right\|^{2}_{\mathbf{P}^{-1}}}\nonumber\\ 
        & + \frac{1}{2} \sum_{k=0}^{L} \left\|\mathbf{y}_{k}-\boldsymbol{\mathcal{H}}_{k}\circ\boldsymbol{\mathcal{M}}^{\mathsf{nn}}_{k:0}\left(\mathbf{p}^{\mathsf{i}}+\delta\mathbf{p}, \mathbf{x}^{\mathsf{i}}_{0}+\delta\mathbf{x}_{0}\right)\right\|^{2}_{\mathbf{R}^{-1}_{k}},\\
        \approx{} &{\frac{1}{2} \left\|\mathbf{x}^{\mathsf{i}}_{0}-\mathbf{x}^{\mathsf{b}}_{0}+\delta\mathbf{x}_{0}\right\|^{2}_{\mathbf{B}^{-1}} + \frac{1}{2} \left\|\mathbf{p}^{\mathsf{i}}-\mathbf{p}^{\mathsf{b}}+\delta\mathbf{p}\right\|^{2}_{\mathbf{P}^{-1}}}\nonumber\\
        & + \frac{1}{2} \sum_{k=0}^{L} \left\|\mathbf{d}_{k}-\mathbf{H}_{k}\mathbf{M}^{\mathsf{nn}}_{k:0}\left(\delta\mathbf{p}, \delta\mathbf{x}_{0}\right)^{\top}\right\|^{2}_{\mathbf{R}^{-1}_{k}},\\
        \triangleq{} & \widehat{\mathcal{J}}^{\mathsf{nn}}\left(\delta\mathbf{p}, \delta\mathbf{x}_{0}\right)\label{eq:methodology-nn4dvar-simplified-incremental-cost}.
    \end{align}
\end{subequations}
where $\mathbf{d}_{k}\triangleq\mathbf{y}_{k}-\boldsymbol{\mathcal{H}}_{k}\circ\boldsymbol{\mathcal{M}}^{\mathsf{nn}}_{k:0}\left(\mathbf{p}^{\mathsf{i}}, \mathbf{x}^{\mathsf{i}}_{0}\right)$, $\mathbf{H}_{k}$ is the tangent linear (TL) operator of $\boldsymbol{\mathcal{H}}_{k}$ taken at $\boldsymbol{\mathcal{M}}^{\mathsf{nn}}_{k:0}\left(\mathbf{p}^{\mathsf{i}}, \mathbf{x}^{\mathsf{i}}_{0}\right)$, and $\mathbf{M}^{\mathsf{nn}}_{k:0}$ is the TL operator of $\boldsymbol{\mathcal{M}}^{\mathsf{nn}}_{k:0}$ taken at $\left(\mathbf{p}^{\mathsf{i}}, \mathbf{x}^{\mathsf{i}}_{0}\right)$. The linearised or \emph{incremental} cost function $\widehat{\mathcal{J}}^{\mathsf{nn}}$ is sometimes also called the \emph{quadratic} cost function because it has the advantage of being quadratic in $\delta\mathbf{p}$ and $\delta\mathbf{x}_{0}$, where the conjugate gradient algorithm could be very efficient. Its gradient can be computed using \cref{alg:grad-generic-incremental-online-augmented-cswrc}, in which the following notation has been used: $\mathbf{F}^{\mathsf{p}}$ and $\mathbf{F}^{\mathsf{x}}$ are the TL operators of $\boldsymbol{\mathcal{F}}$ with respect to $\mathbf{p}$ and $\mathbf{x}$, respectively, both taken at $\left(\mathbf{p}^{\mathsf{i}}, \mathbf{x}^{\mathsf{i}}_{0}\right)$, and $\mathbf{M}_{k+1:k}$ is the TL operator of $\boldsymbol{\mathcal{M}}_{k+1:k}$ taken at $\boldsymbol{\mathcal{M}}_{k:0}\left(\mathbf{p}^{\mathsf{i}}, \mathbf{x}^{\mathsf{i}}_{0}\right)$. In this algorithm, \cref{step:grad-generic-incremental-online-augmented-cost-start-wc4dvar,step:i1,step:i2,step:i3,step:i4,step:i5,step:i6,step:i7,step:i8,step:i9,step:i10,step:i11,step:grad-generic-incremental-online-augmented-cost-end-wc4dvar} corresponds to the gradient of the incremental cost function of the WC 4D-Var cost function (without the background terms).

\begin{algorithm}[tbp]
    \caption{\label{alg:grad-generic-incremental-online-augmented-cswrc}Gradient of the incremental cost function $\widehat{\mathcal{J}}^{\mathsf{nn}}$ \cref{eq:methodology-nn4dvar-simplified-incremental-cost}.}
    \algnewcommand\algorithmicto{\textbf{to}}
    \newcommand{\algorithmicoutput}{\textbf{Output:}}
    \newcommand{\Output}{\item[\algorithmicoutput]}
    \begin{algorithmic}[1]
    \renewcommand{\algorithmicensure}{\textbf{Input:}}
    \Ensure{$\delta\mathbf{p}$ and $\delta\mathbf{x}_{0}$}
    \State{\label{step:grad-generic-incremental-online-augmented-cost-tl-nn}$\delta\mathbf{w}\gets\mathbf{F}^{\mathsf{p}}\delta\mathbf{p}+\mathbf{F}^{\mathsf{x}}\delta\mathbf{x}_{0}$}\Comment{TL of the NN $\boldsymbol{\mathcal{F}}$}
    \State{\label{step:grad-generic-incremental-online-augmented-cost-start-wc4dvar}$\mathbf{z}_{0}\gets\mathbf{R}^{-1}_{0}\left(\mathbf{H}_{0}\delta\mathbf{x}_{0}-\mathbf{d}_{0}\right)$}
    \For{\label{step:i1}$k=1$ \algorithmicto~$L-1$}
        \State{\label{step:i2}$\delta\mathbf{x}_{k}\gets\mathbf{M}_{k:k-1}\delta\mathbf{x}_{k-1}+\delta\mathbf{w}$}\Comment{TL of the dynamical model $\boldsymbol{\mathcal{M}}_{k:k-1}$}
        \State{\label{step:i3}$\mathbf{z}_{k}\gets\mathbf{R}^{-1}_{k}\left(\mathbf{H}_{k}\delta\mathbf{x}_{k}-\mathbf{d}_{k}\right)$}
    \EndFor\label{step:i4}
    \State{\label{step:i5}$\delta\tilde{\mathbf{x}}_{L-1}\gets\mathbf{0}$}\Comment{AD variable for system state}
    \State{\label{step:i6}$\delta\tilde{\mathbf{w}}_{L-1}\gets\mathbf{0}$}\Comment{AD variable for model error}
    \For{\label{step:i7}$k=L-1$ \algorithmicto~$1$}
        \State{\label{step:i8}$\delta\tilde{\mathbf{x}}_{k}\gets\mathbf{H}^{\top}_{k}\mathbf{z}_{k}+\delta\tilde{\mathbf{x}}_{k}$}
        \State{\label{step:i9}$\delta\tilde{\mathbf{w}}_{k-1}\gets\delta\tilde{\mathbf{x}}_{k}+\delta\tilde{\mathbf{w}}_{k}$}
        \State{\label{step:i10}$\delta\tilde{\mathbf{x}}_{k-1}\gets\mathbf{M}^{\top}_{k:k-1}\delta\tilde{\mathbf{x}}_{k}$}\Comment{AD of the dynamical model $\boldsymbol{\mathcal{M}}_{k:k-1}$}
    \EndFor\label{step:i11}
    \State{\label{step:grad-generic-incremental-online-augmented-cost-end-wc4dvar}$\delta\tilde{\mathbf{x}}_{0}\gets\mathbf{H}^{\top}_{0}\mathbf{z}_{0}+\delta\tilde{\mathbf{x}}_{0}$}
    \State{\label{step:grad-generic-incremental-online-augmented-cost-adx-nn}$\delta\tilde{\mathbf{x}}_{0}\gets\left[\mathbf{F}^{\mathsf{x}}\right]^{\top}\delta\tilde{\mathbf{x}}_{0}$}\Comment{AD of the NN $\boldsymbol{\mathcal{F}}$}
    \State{\label{step:grad-generic-incremental-online-augmented-cost-adp-nn}$\delta\tilde{\mathbf{p}}\gets\left[\mathbf{F}^{\mathsf{p}}\right]^{\top}\delta\tilde{\mathbf{w}}_{0}$}\Comment{AD of the NN $\boldsymbol{\mathcal{F}}$}
    \State{$\delta\tilde{\mathbf{x}}_{0}\gets\mathbf{B}^{-1}\left(\mathbf{x}^{\mathsf{i}}_{0}-\mathbf{x}^{\mathsf{b}}_{0}+\delta\mathbf{x}_{0}\right)+\delta\tilde{\mathbf{x}}_{0}$}
    \State{$\delta\tilde{\mathbf{p}}\gets\mathbf{P}^{-1}\left(\mathbf{p}^{\mathsf{i}}-\mathbf{p}^{\mathsf{b}}+\delta\mathbf{p}\right)+\delta\tilde{\mathbf{p}}$}
    \Output{$\nabla_{\delta\mathbf{p}}\widehat{\mathcal{J}}^{\mathsf{nn}}=\delta\tilde{\mathbf{p}}$ and $\nabla_{\delta\mathbf{x}_{0}}\widehat{\mathcal{J}}^{\mathsf{nn}}=\delta\tilde{\mathbf{x}}_{0}$}
    \end{algorithmic}
\end{algorithm}

In the end, in order to implement the simplified NN 4D-Var we can reuse most of the framework already in place for WC 4D-Var and we need to provide:
\begin{itemize}
    \item the forward operator $\boldsymbol{\mathcal{F}}$ of the NN to compute the nonlinear trajectory at the start of each outer iteration;
    \item the TL operators $\mathbf{F}^{\mathsf{x}}$ and $\mathbf{F}^{\mathsf{p}}$ of the NN for \cref{step:grad-generic-incremental-online-augmented-cost-tl-nn} of \cref{alg:grad-generic-incremental-online-augmented-cswrc};
    \item the adjoint (AD) operators $\left[\mathbf{F}^{\mathsf{x}}\right]^{\top}$ and $\left[\mathbf{F}^{\mathsf{p}}\right]^{\top}$ of the NN for
    \cref{step:grad-generic-incremental-online-augmented-cost-adx-nn,step:grad-generic-incremental-online-augmented-cost-adp-nn} of \cref{alg:grad-generic-incremental-online-augmented-cswrc}.
\end{itemize}
From a technical perspective, all these operators have to be computed in the model core, where the components of the system state are available. In OOPS, the model core is implemented in Fortran, which implies that we need a ML library in Fortran. The only one that we could find, namely the Fortran--Keras Bridge \citep[FKB,][]{ott-2020}, does not provide all the required operators. For this reason, we have implemented our own NN library in Fortran, called Fortran neural networks \citep[FNN,][]{fnn-2022}. In this library, we have manually implemented, for each layer that we need, functions for the forward, but also the TL and adjoint operators with respect to both NN parameters and NN input. We have then included the FNN library in OOPS and added the interface between OOPS and FNN for two forecast models, OOPS-QG and OOPS-IFS. Finally, we have included the NN parameters in the control variables in OOPS, in such a way that they can be estimated using the simplified NN 4D-Var method.

\section{The quasi-geostrophic model}
\label{sec:qg}

The simplified NN 4D-Var formulation provides a convenient alternative to the original NN 4D-Var. It has the advantage of being much easier to implement because it is built on top of WC 4D-Var, which is already implemented in OOPS. We first test and validate the method using OOPS-QG. In particular, we want to confirm that the simplified NN 4D-Var method is able to make an accurate online estimation of model error.

\subsection{Brief model description}

The quasi-geostrophic (QG) model in the present article is the same as the one used by \citet{fisher-2017, laloyaux-2020} and later by \citet{farchi-2021}. In the following, we only outline the model description. More details about this model can be found in \citet{fisher-2017, laloyaux-2020}.

The QG model's equations express the conservation of the (non-dimensional) potential vorticity $q$ for two layers of constant potential temperature in the $x-y$ plane. The potential vorticity is related to the stream function $\psi$ through a specific variant of Poisson's equation. The domain is periodic in the $x$ direction, and with fixed boundary conditions for $q$ in the $y$ direction. We use a horizontal discretisation of $\num{40}$ grid points in the $x$ direction and $\num{20}$ in the $y$ direction. In OOPS, the control vector $\mathbf{x}$ contains all values of the stream function $\psi$ for both levels, \ie{} a total of $N_{\mathsf{x}}=\num{1600}$ variables.

\subsection{The reference and perturbed setups}

In the test series reported in \cref{sec:offline,sec:online}, we rely on twin experiments. The synthetic truth is generated using the reference setup described by \citet{farchi-2021}. Model error is then introduced by using a perturbed setup, in which the values of both layer depths and the integration time steps have been modified, as reported in \cref{tab:qg-parameter-perturbation}. Note that, by contrast with the perturbed setup of \citet{farchi-2021}, the orography term has not been changed, because we have found that the model error setup is sufficiently challenging as is and an orography perturbation does not add meaningful complexity here.

\begin{table}[tbp]
    \centering
    \caption{\label{tab:qg-parameter-perturbation}Set of parameters for the reference setup (middle row) and the perturbed setup (right row).}
    \begin{tabular}{lrr}
    \toprule
    Parameter & Reference setup & Perturbed setup \\
    \midrule
    Top layer depth & \SI{6000}{\meter} & \SI{5750}{\meter} \\
    Bottom layer depth & \SI{4000}{\meter} & \SI{4250}{\meter} \\
    Integration time step & \SI{10}{\min} & \SI{20}{\min} \\
    \bottomrule 
    \end{tabular}
\end{table}

\subsection{Neural network architecture for model error correction}
\label{ssec:qg-nn-archiecture}

By construction, NN 4D-Var (both the original and simplified formulations) is very similar to parameter estimation, which is very challenging when the number of parameters is high. For this reason, it is important to use smart NN architectures to be \emph{parameter efficient}, \ie{} reduce as much as possible the number of parameters. This typically involves applying prior knowledge about the system under study to the choice of the NN architecture. A typical smart architecture is the monomial architecture introduced by \citet{bocquet-2019a}, in which the model tendencies are parametrised by a set of regressors (the monomials) and then integrated in time to build the resolvent between two time steps. In the present article, we follow another approach, introduced by \citet{bonavita-2020} for the IFS. In this case, the NN is applied independently for each atmospheric column and for several groups of variables: mass (temperature and surface pressure), wind (vorticity and divergence), and humidity. Horizontal and temporal variations are taken into account by adding latitude, longitude, time of the day, and month of the year to the set of predictors. This choice is imposed by operational constraints -- variables in different columns may come from different processes when using parallelism. It also makes sense because a significant amount of the model error in the IFS comes from the parametrisation of physical processes, which is applied in vertical model columns \citep{polichtchouk-2022}, and because in this configuration, the amount of samples is multiplied by the number of vertical columns in the data, which is highly beneficial to the training. Furthermore, it has been shown that the performance of simple vertical NNs is roughly similar to that of non-vertical convolutional neural networks in a realistic model error correction problem \citep{laloyaux-2022}.

\begin{figure}[tbp]
    \centering
    \includegraphics[scale=1]{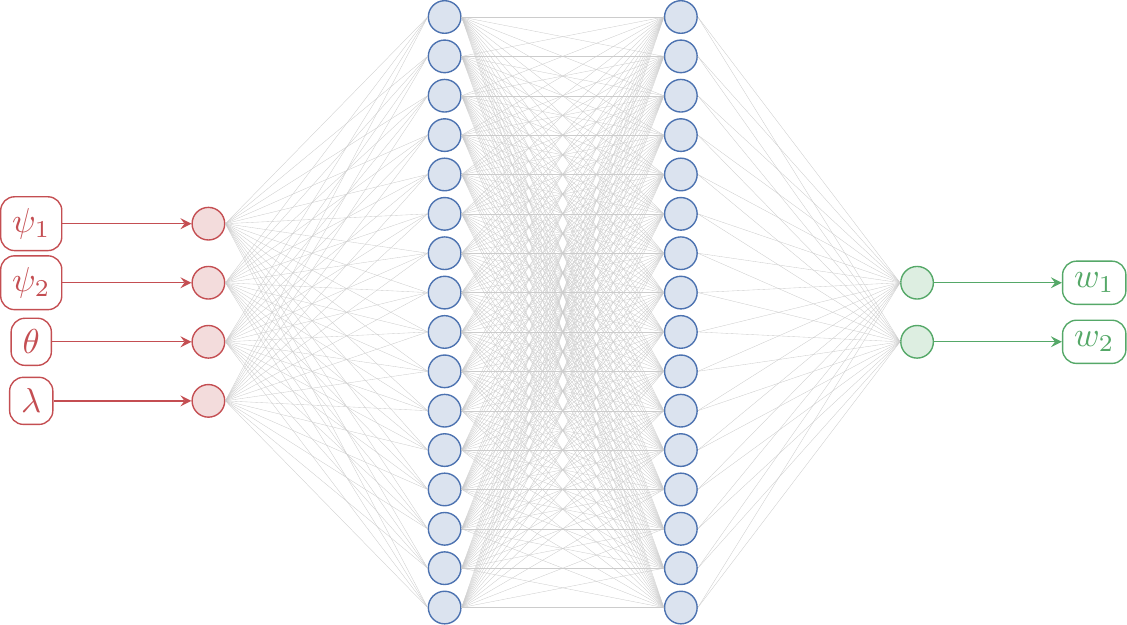}
    \caption{\label{fig:qg-nn-architecture}Illustration of the NN architecture. On the left in red, the input layer. In the centre in blue, the two hidden layers, with $\tanh$ activation. On the right in green, the output layer.}
\end{figure}

The QG model has only two vertical layers and one variable, the stream function $\psi$, and it is autonomous, \ie{} the model does not explicitly depend on time. This means that our NN for model error correction, independently applied to all $\num{40}\times\num{20}$ columns, has four predictors:
\begin{enumerate}
    \item $\psi_{1}$ the bottom layer stream function;
    \item $\psi_{2}$ the top layer stream function;
    \item $\sin\left[2\pi\left(\theta-1/2\right)/40\right]$, where $\theta$ is the longitude index between $\numlist{1; 40}$;
    \item $\sin\left[\pi\left(\lambda-1/2-10\right)/20\right]$
    where $\lambda$ is the latitude index between $\numlist{1; 20}$;
\end{enumerate}
and two predictands:
\begin{enumerate}
    \item $w_{1}$ the model error estimate for the bottom layer stream function;
    \item $w_{2}$ the model error estimate for the top layer stream function.
\end{enumerate}
Note that the sinus function is used here to make the NN aware of the periodicity. We have tested several NNs, and ended up with the following sequential architecture, illustrated in \cref{fig:qg-nn-architecture}: (i) a first internal dense layers with $\num{16}$ neurons and with the $\tanh$ activation function; (ii) 
a second internal dense layers with $\num{16}$ units and with the $\tanh$ activation function as well; (iii) one output dense layer with $\num{2}$ units and no activation function. This NN has a total of $\left(\num{2}\times\num{4} + \num{4}\right) + \left(\num{4}\times\num{4} + \num{4}\right) + \left(\num{4}\times\num{2} + \num{2}\right)=\num{386}$ parameters, which is significantly less than the number of variables ($\num{1600}$).

To stay within the scope of the simplified NN 4D-Var defined in \cref{ssec:methodology-nn4dvar-simplified}, we assume that the NN correction is constant throughout the window, and that it is added after every model time step (\ie{} every $\SI{20}{\min}$ in our case) as it is enforced in the current implementation of WC 4D-Var. According to the classification of \citet{farchi-2021, farchi-2021b}, this approach is a \emph{resolvent} correction, because it is added after the integration scheme. However, a classical resolvent correction would add the correction after every window, in other words much less frequently than after every model time step. Hence, the spirit of the present correction is closer to that of a \emph{tendency} correction.

\section{Offline learning results}
\label{sec:offline}

We begin the numerical experiments by using offline learning to train the NN. Offline learning here serves two purposes: it provides a baseline for comparison as well as a pre-trained NN for online learning.

\subsection{Observation and data assimilation setup}
\label{ssec:offline-obs-da-setup}

In the present test series, we use for the QG model the same initial condition as \citet{farchi-2021}. After a first relaxation run of $\SI{256}{\day}$, the state is perturbed and a second relaxation run of $\SI{256}{\day}$ is performed to provide the initial state for the DA experiment. At this point, observations are available every $\SI{2}{\hour}$, starting at 01:00 every day, at $\num{30}$ fixed locations, whose distribution mimics the coverage provided by (polar-orbiting) satellite soundings. The observation operator is simply a bilinear interpolation of the stream function at the observation locations. The observations are independently perturbed using a Gaussian noise with zero mean and standard deviation equal to $\num{0.2}$ (about $\SI{4}{\percent}$ of the model variability).

We start by assimilating the observations using cycled strong-constraint 4D-Var, with consecutive windows of $\SI{1}{\day}$ starting at 00:00 each. Hence, there are $\num{12}$ batches of observations, for a total of $\num{360}$ observations per window. The observation error covariance matrix is set to $\mathbf{R}=\num{0.2}^{2}\mathbf{I}$ to be consistent with how the synthetic observations are produced. For the first cycle, the background state $\mathbf{x}^{\mathsf{b}}_{0}$ is set to be the initial condition before the two relaxation runs. For the following cycles, the background state is obtained by forecasting the previous analysis state. Finally, the background error covariance matrix is set to $\mathbf{B}=b^{2}\mathbf{C}$, where $\mathbf{C}$ is a short-range correlation matrix, the same as the one used by \citet{farchi-2021}, and where $b$ is the standard deviation, a free parameter. The accuracy of the estimations is measured with the instantaneous root-mean-squared error (RMSE) between the estimate and the truth for all $\num{1600}$ state variables, possibly averaged over time. In particular, the first-guess (respectively analysis) RMSE is defined in this article as the instantaneous RMSE between the first-guess (or analysis) trajectory, the trajectory originated from the first-guess (or analysis) at the start of the window, and the true trajectory, averaged over the entire DA window. The time-averaged first-guess (respectively analysis) RMSE is then defined as this first-guess (or analysis) RMSE averaged over a sufficiently large number of cycles.

In order to be close to operational conditions, we tune the value of $b$ to minimise the time-averaged first-guess RMSE. Preliminary experiments (not detailed here) have shown that, for the present DA setup, the optimal value is $b=\num{0.4}$. With this value, we run a cycled DA experiment of $N^{\text{total}}_{\mathsf{t}}=\num{2100}$ cycles. The results of the first $N^{\text{spinup}}_{\mathsf{t}}=\num{51}$ cycles are dropped as spin-up process of the experiment. Then, for each remaining cycles $t=\num{1}, \ldots, N^{\text{data}}_{\mathsf{t}}=\num{2049}$, we keep $\mathbf{x}^{\mathsf{b}}_{0}\left(t\right)$ and $\mathbf{x}^{\mathsf{a}}_{0}\left(t\right)$, respectively the first-guess and the analysis at the start of the $t$-th window.

\subsection{Neural network training}
\label{ssec:offline-nn-training}

As shown by \citet{farchi-2021}, the analysis increment $\mathbf{x}^{\mathsf{a}}_{0}\left(t\right)-\mathbf{x}^{\mathsf{b}}_{0}\left(t\right)$ can be chosen as a proxy of the model error for a $\num{1}$-window-long integration, provided that the analysis is a reasonably accurate estimation of the true state:
\begin{equation}
    \label{eq:offline-nn-training-analysis-incr}
    \mathbf{x}^{\mathsf{a}}_{0}\left(t+1\right)-\mathbf{x}^{\mathsf{b}}_{0}\left(t+1\right) = \mathbf{x}^{\mathsf{a}}_{0}\left(t+1\right) - \boldsymbol{\mathcal{M}}_{t}\left(\mathbf{x}^{\mathsf{a}}_{0}\left(t\right)\right) \approx \mathbf{x}^{\mathsf{t}}_{0}\left(t+1\right) - \boldsymbol{\mathcal{M}}_{t}\left(\mathbf{x}^{\mathsf{t}}_{0}\left(t\right)\right),
\end{equation}
where $\boldsymbol{\mathcal{M}}_{t}$ corresponds to the resolvent of the model between the start of the $t$-th window and the start of the $(t+1)$-th window, and where $\mathbf{x}^{\mathsf{t}}_{0}\left(t\right)$ is the true state of the system at the start of the $t$-th window. However, as explained in \cref{ssec:qg-nn-archiecture}, the NN correction is added after every model time step, which means that we need a proxy of the model error for a $\num{1}$-step integration. Without further knowledge on the model error dynamics, we assume a uniform linear growth of model error in time and hence we rescale the analysis increments by a factor $\delta t/\Delta T=\num{1}/\num{72}$, where $\delta t=\SI{20}{\min}$ is the model time step and $\Delta T=\SI{1}{\day}$ is the window length. Note that, even if the analysis was available at a $\num{1}$ model step frequency, we would not use it because the accuracy of the analysis would most probably be insufficient to detect a model error signal in the analysis increments.

To summarise, we use the following dataset for the training of the NN:
\begin{equation}
    \left\{\mathbf{x}^{\mathsf{a}}_{0}\left(t\right)\mapsto\frac{\delta t}{\Delta t}\left[\mathbf{x}^{\mathsf{a}}_{0}\left(t+1\right)-\mathbf{x}^{\mathsf{b}}_{0}\left(t+1\right)\right], \quad t=\num{1}, \ldots, N^{\text{data}}_{\mathsf{t}}-1=\num{2048}\right\}.
\end{equation}
Note the time lag between the input $\mathbf{x}^{\mathsf{a}}_{0}\left(t\right)$ and the output $\delta t/\Delta T\left(\mathbf{x}^{\mathsf{a}}_{0}\left(t+1\right)-\mathbf{x}^{\mathsf{b}}_{0}\left(t+1\right)\right)$. Indeed, the analysis increment $\mathbf{x}^{\mathsf{a}}_{0}\left(t+1\right)-\mathbf{x}^{\mathsf{b}}_{0}\left(t+1\right)$ of the $(t+1)$-th window does inform about the model error during the $t$-th window, which is exactly what we need according to the model formulation described in \cref{ssec:methodology-nn4dvar-simplified}. Also note that we have chosen to use the analysis $\mathbf{x}^{\mathsf{a}}_{0}\left(t\right)$ as predictor, but we could have equivalently chosen the first-guess $\mathbf{x}^{\mathsf{b}}_{0}\left(t\right)$. Preliminary experiments (not illustrated here) have shown that both choices yield similar results. Since the NN is applied independently to each atmospheric column, there are actually $\num{40}\times\num{20}=\num{800}$ samples per pair (analysis $\mapsto$ analysis increment). Finally, in order to accelerate the convergence, the input and output of the training dataset are standardised before the training, using independent normalisation coefficients for each variable.

In order to evaluate the sensitivity to the length of the dataset, we train the NN using only the last $N^{\text{train}}_{\mathsf{t}}$ pairs (analysis $\mapsto$ analysis increment) for several values of $N_{\mathsf{t}}$. Among all these $N^{\text{train}}_{\mathsf{t}}$ pairs, the first $7/8$\textsuperscript{th} form the training dataset and the last $1/8$\textsuperscript{th} the validation dataset. The test dataset is formed by $N^{\text{test}}_{\mathsf{t}}=\num{2048}$ pairs (truth $\mapsto$ true model error) originated from a different trajectory of the model. With this setup, the NN is trained for a maximum of $\num{1024}$ epochs using Adam algorithm \citep{kingma-2015}, a variant of the stochastic gradient descent, with the typical learning rate $\num{1e-3}$. The loss function is the mean-squared error (MSE). To accelerate the training, we use a relatively large batch size ($\num{1024}$) as well as an early stopping callback on the validation MSE with a patience of $\num{256}$ epochs. After the training, we compute the test MSE. This experiment is repeated $\num{16}$ times with different sets of trajectories for training and testing and different random seeds for Adam. For comparison, we have also performed the exact same set of experiments but with dense and perfect observations, \ie{} when the analysis is equal to the true state. This second set of experiments illustrates the full predictive power of the NN representation of the model error.

\begin{figure}[tbp]
    \centering
    \includegraphics[scale=1]{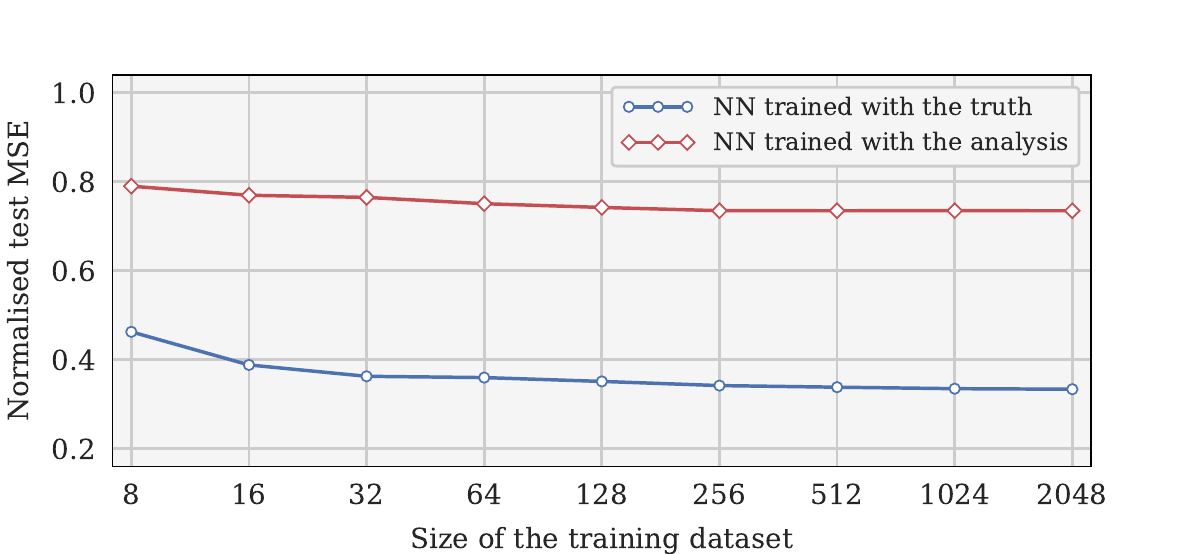}
    \caption{\label{fig:offline-nn-training}Offline NN training. Evolution of the normalised test MSE as a function of the length of the training dataset $N^{\text{train}}_{\mathsf{t}}$ for the NN trained with the truth (in blue) and the NN trained with the analysis (in red).}
\end{figure}

\Cref{fig:offline-nn-training} shows the evolution of the test MSE as a function of the length of the training dataset $N^{\text{train}}_{\mathsf{t}}$. The score is normalised by the averaged squared norm of the model error, in such a way that it is equal to $\num{1}$ when the NN predicts a zero model error. In all experiments, the normalised test MSE is lower than $\num{1}$. This means that, on average, the model error prediction is useful. When using the truth, both training and test datasets are statistically equivalent. The normalised test MSE decreases with the size of the training dataset $N^{\text{train}}_{\mathsf{t}}$. The final value is $\num[round-mode=places, round-precision=3]{0.33361095}$ for $N^{\text{train}}_{\mathsf{t}}=\num{2048}$, but the score is already quite good ($\num[round-mode=places, round-precision=3]{0.35116949}$) for $N^{\text{train}}_{\mathsf{t}}=\num{128}$. The residual error for a large training dataset comes from the limited predictive power of the NN. We have checked that better scores can easily be obtained when using larger, non column-wise NNs. Unsurprisingly, when using the analysis the normalised test MSE is significantly higher ($\num[round-mode=places, round-precision=3]{0.73481396}$ at best) and stops improving for $N^{\text{train}}_{\mathsf{t}}\geq\num{256}$. The primary reason for these discrepancies is the fact that the statistical moments (\eg{} the time average and time standard deviation) are not the same between the analysis increments and the true model error. In particular, the average analysis increment norm is lower than the average model error norm. This means that the NN trained with the analysis generally underestimates the model error. This is consistent with what has been found by \citet{crawford-2020, farchi-2021}.

\subsection{Corrected data assimilation}
\label{ssec:offline-corrected-da}

Now that the NN has been trained, we would like to test the hybrid model in forecast and DA experiments. We start with DA using the exact same setup as in \cref{ssec:offline-obs-da-setup}, but with a true state taken from a different trajectory of the model. Four 4D-Var variants are compared:
\begin{enumerate}
    \item \labelSC{}: strong-constraint with the physical model (no model error correction).
    \item \labelWC{}: weak-constraint with the physical model -- in this case the model error correction comes from the constant, online estimated forcing.
    \item \labelSCNNt{}: strong-constraint with the hybrid model, where the NN correction has been trained with the truth using the largest dataset ($N^{\text{train}}_{\mathsf{t}}=\num{2048}$).
    \item \labelSCNNa{}: strong-constraint with the hybrid model, where the NN correction has been trained with the analysis using the largest dataset ($N^{\text{train}}_{\mathsf{t}}=\num{2048}$).
\end{enumerate}
In all cases, we use the same background error covariance matrix $\mathbf{B}$ as in \cref{ssec:offline-obs-da-setup}, because we want to highlight the benefit of each approach without the need to re-tune $\mathbf{B}$. The initial background state $\mathbf{x}^{\mathsf{b}}_{0}$ corresponds to the background obtained after a spin-up of $\num{32}$ DA cycles with strong-constraint 4D-Var. For weak-constraint 4D-Var, we need to provide in addition (i) the initial background for model error $\mathbf{w}^{\mathsf{b}}\left(0\right)$, and (ii) the background error covariance matrix for model error $\mathbf{Q}$. We choose to use $\mathbf{w}^{\mathsf{b}}\left(0\right)=\mathbf{0}$ and $\mathbf{Q}=q^{2}\widehat{\mathbf{C}}$, where $\widehat{\mathbf{C}}$ is a long-range correlation matrix, the same as the one used by \citet{laloyaux-2020}, and where $q$ is the standard deviation, another free parameter. We choose $q=\num{0.004}$ in order to minimise the time-averaged first-guess RMSE. In each case, we run a cycled DA experiment of $N^{\text{assim}}_{\mathsf{t}}=\num{257}$ cycles, which we empirically consider to be sufficiently long. The results of the first $\num{33}$ cycles are dropped as spin-up. For the remaining $\num{224}$ cycles, we compute the first-guess and analysis RMSE. Each experiment is repeated $\num{128}$ times with different trajectories for the synthetic truth. Note that in the second and third case, the $\num{128}$ repetitions are equally spread over the $\num{16}$ trained NN obtained in \cref{ssec:offline-nn-training}: experiments $\numrange{1}{8}$ use the first trained NN, experiments $\numrange{9}{16}$ use the second, experiments $\numrange{17}{24}$ use the third, etc.

\begin{table}[tbp]
    \centering
    \caption{\label{tab:offline-da-results}Offline DA results. Time-averaged first-guess and analysis RMSE for the four 4D-Var variants presented in \cref{ssec:offline-corrected-da}. For each variant, we report the mean (main numbers) and standard deviation (in parentheses) values over the $\num{128}$ experiments.}
    \sisetup{round-mode=places, round-precision=3}
    \begin{tabular}{lllrr}
    \toprule
    Variant & 4D-Var constraint & Model error correction & First-guess RMSE & Analysis RMSE \\
    \midrule
    \labelSC{} & strong & --- & \num{0.34970503987615065} (\num{0.019995550888214204}) & \num{0.15676421646753297} (\num{0.0031826990857734736}) \\
    \labelWC{} & weak & constant, online estimated & \num{0.2713846038523295} (\num{0.015541281716700507}) & \num{0.1278801870723817} (\num{0.0029347220600346342}) \\
    \labelSCNNt{} & strong & NN trained offline with the truth & \num{0.2634069931666347} (\num{0.01841594453111139}) & \num{0.1325750575183806} (\num{0.0025712425642218832}) \\
    \labelSCNNa{} & strong & NN trained offline with the analysis & \num{0.26523341750323143} (\num{0.02021157964043708}) & \num{0.1438638981222866} (\num{0.002746151861755436}) \\
    \bottomrule 
    \end{tabular}
\end{table}

The time-averaged first-guess and analysis RMSE are reported in \cref{tab:offline-da-results}. The results show the efficiency of model error corrections: in all cases, the first-guess and the analysis are more accurate with model error correction (\labelWC{}/\labelSCNNt{}/\labelSCNNa{}) than without (\labelSC{}). As expected, the model error correction provided by the NN is more efficient when the NN has been trained with the truth (\labelSCNNt{}) than when it has been trained with the analysis (\labelSCNNa{}). Furthermore, using the offline correction provided by the NN (\labelSCNNt{}/\labelSCNNa{}) yields in both cases a more accurate first-guess but a less accurate analysis than using the online correction computed with weak-constraint 4D-Var (\labelWC{}).

\subsection{Corrected forecast}
\label{ssec:offline-corrected-forecast}

\begin{figure}[tbp]
    \centering
    \includegraphics[scale=1]{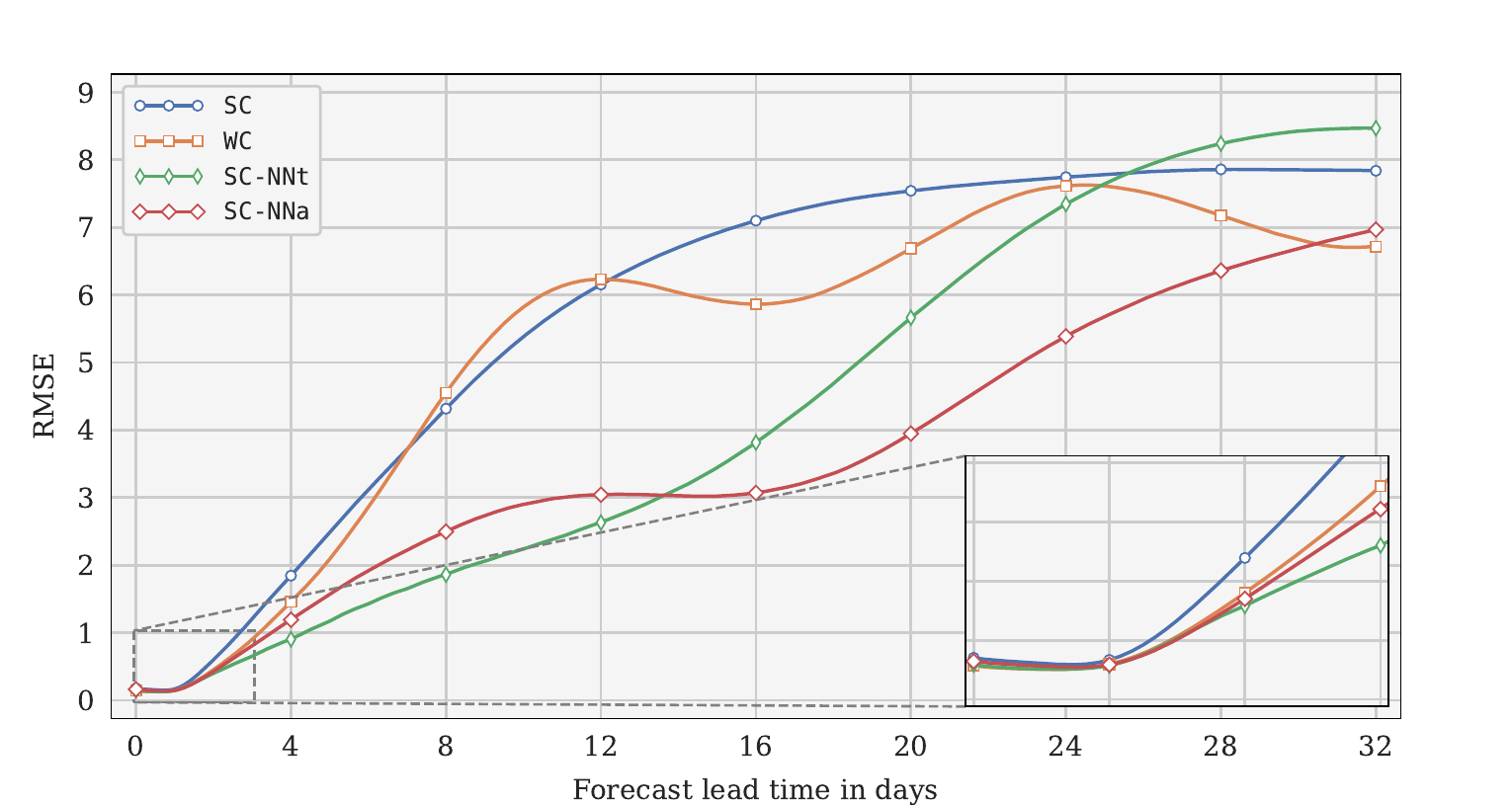}
    \caption{\label{fig:offline-corrected-forecast}Offline forecast results. Evolution of the forecast RMSE, averaged over the last $\num{32}$ cycles and over the $\num{128}$ experiments, as a function of the forecast lead time for the four 4D-Var variants: \labelSC{} in blue, \labelWC{} in orange, \labelSCNNt{} in green, and \labelSCNNa{} in red. The insert zooms in the short forecast lead times.}
\end{figure}

To conclude this first test series, we evaluate the accuracy of the model in the four cases described in \cref{ssec:offline-corrected-da}. To this end, we extend the previous set of experiments. After each analysis cycle, we compute a $\num{32}$-day forecast starting from the DA analysis using the same model as in the 4D-Var cost function. In the case of weak-constraint 4D-Var (\labelWC{}), the constant, online estimated forcing is used throughout the entire forecast. In the case of strong-constraint 4D-Var with the hybrid model (\labelSCNNt{}/\labelSCNNa{}), the NN correction is also used throughout the entire forecast, but in a flow-dependent way: the correction values are updated at a $\num{1}$-day frequency using the forecasted state. With these specifications, the error in the first day of forecast corresponds to the analysis error and the error in the second day of forecast corresponds to the first-guess error. \Cref{fig:offline-corrected-forecast} shows the evolution of the forecast RMSE, averaged over the last $\num{32}$ DA cycles and over the $\num{128}$ repetitions of the experiments, as a function of the forecast lead time.

With weak-constraint 4D-Var (\labelWC{}), the model error correction is calibrated over the DA window, i.e. over the first day. Overall, the correction is efficient and yields a more accurate forecast than with the non-corrected model (\labelSC{}). After several days, the true model error has significantly evolved and this initial error estimate gets less accurate. This is why the reduction of the forecast error vanishes after several days. Also note that the model has a periodic behaviour, with a period around $\num{16}$ days. This means that, after $\num{16}$ days, the model state (and hence the model error) is roughly the same as at the beginning, which explains the forecast error reduction around day $\num{16}$ and around day $\num{32}$.

By contrast, when using the hybrid model (\labelSCNNt{}/\labelSCNNa{}), the model error correction is flow-dependent (updated every day). This yields overall an even more accurate forecast than with weak-constraint 4D-Var (\labelWC{}). In the first few days, the correction accumulates and positively interacts with the physical model, which is why the forecast error reduction increases over time. After several days however, the model error correction becomes less efficient, because the forecasted state -- the most important predictor of the NN -- has become significantly different from the true state. At this point, the model error correction does not any more yield a forecast error reduction. Worse, it even increases the forecast errors. This explains the quick increase of the forecast errors after $\num{10}$ days when the NN is trained with the truth (\labelSCNNt{}) and after $\num{15}$ days when the NN is trained with the analysis (\labelSCNNa{}). In an operational perspective, it would be interesting to progressively mitigate the model error correction over time, but this is beyond the scope of the present study. Surprisingly, the validity period of the model error correction is longer for \labelSCNNa{} (NN trained with the analysis) than for \labelSCNNt{} (NN trained with the truth). This could be due to the fact that a NN trained with the analysis underestimates the model error: if the model error estimate is pointing in the wrong direction, it is better to have an underestimated model error \citep{crawford-2020}. Finally, after about $\num{13}$ days, the forecast is more accurate with \labelSCNNa{}. We believe that this result is related to the limited predictive power of the chosen NN. Indeed, we have checked that with larger NNs, the accuracy of the forecast is always more accurate with \labelSCNNt{} than with \labelSCNNa{}.

\section{Online learning results}
\label{sec:online}

In the present section, we test the simplified online NN 4D-Var presented in \cref{ssec:methodology-nn4dvar-simplified} using the same QG model as in the offline experiments.

\subsection{Data assimilation setup}

In this last test series, we use the same DA setup as in \cref{ssec:offline-obs-da-setup,ssec:offline-corrected-da}. Once again, the true state stems from a different trajectory. We keep the same initial background state $\mathbf{x}^{\mathsf{b}}_{0}$ and background error covariance matrix $\mathbf{B}$ as in \cref{ssec:offline-corrected-da}, once again to highlight the benefit of each approach without the need to re-tune $\mathbf{B}$. In addition, we need to provide (i) the initial background for model parameters $\mathbf{p}^{\mathsf{b}}_{0}$ and (ii) the background error covariance matrix for model parameters $\mathbf{P}$. For $\mathbf{p}^{\mathsf{b}}_{0}$, we choose to use the parameters of the NN that has been trained offline with the analysis, in other words we use offline learning as a pre-training step for online learning. Hence we hope to immediately see the potential benefits of online learning. Finally, without any prior knowledge on the model parameters, we use $\mathbf{P}=p^{2}\mathbf{I}$, where $p$ is the standard deviation, a free parameter. After several preliminary tests, we have chosen $p=\num{0.02}$. Following the approach of \cref{ssec:offline-corrected-forecast}, at each DA cycle, we compute a $\num{32}$-day forecast starting from the DA analysis using the hybrid model with the updated parameters. Finally, once again, each experiment is repeated $\num{128}$ times with as many different trajectories for the synthetic truth. In the following paragraphs, we use the label \labelNN{} to refer to this fifth 4D-Var variant.

\subsection{Temporal evolution of the forecast errors}
\label{ssec:online-forecast-error-temporal-evolution}

\begin{figure}[tbp]
    \centering
    \includegraphics{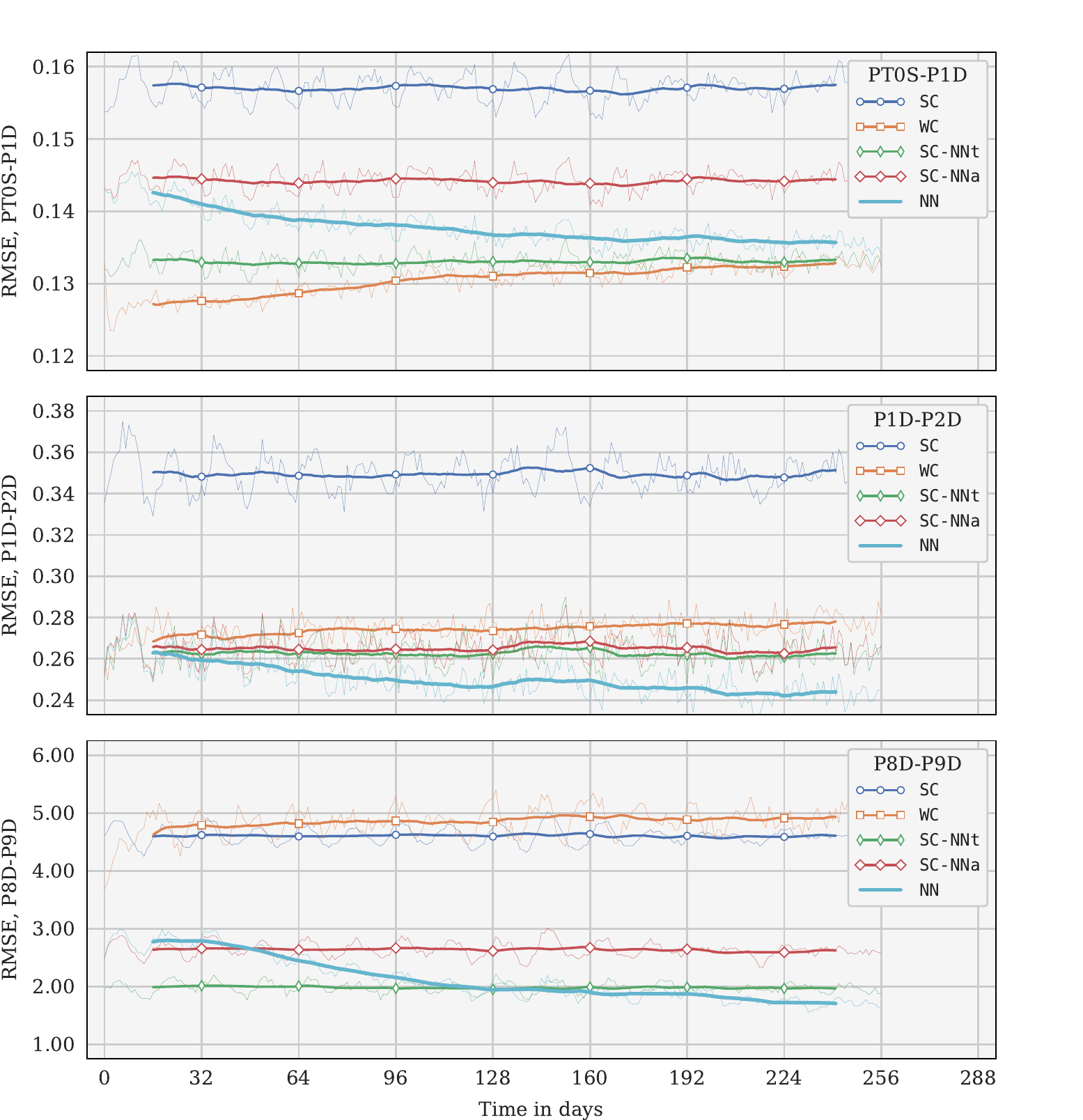}
    \caption{\label{fig:online-rmse-nn} Forecast scores for the online experiments. Evolution of the forecast RMSE, averaged over the $\num{128}$ experiments and over PT0S-P1D (top panel), over P1D-P2D (middle panel), or over P8D-P10D (bottom panel), as a function of time for the five 4D-Var variants: \labelSC{} in blue, \labelWC{} in orange, \labelSCNNt{} in green, \labelSCNNa{} in red, and \labelNN{} in teal. The thin lines report the instantaneous values and the thick lines report the running-average over $\num{32}$ cycles.}
\end{figure}

\Cref{fig:online-rmse-nn} shows the temporal evolution of the errors in the first day of forecast (the analysis), in the second day of forecast (the first-guess), and in the eighth day of forecast (which corresponds to a medium-range forecast). The evolution in all three cases is very similar. At the start of the experiment, the forecast errors with \labelNN{} (NN trained online) are close to those with \labelSCNNa{} (NN trained offline with the analysis). This was expected because in the \labelNN{} variant, we have initialised the parameters of the NN using the parameters obtained by offline training with the analysis. The added positive effect of the online NN training is then rapidly visible. After a few cycles, the forecast errors have decreased. This improvement is quicker for shorter forecast horizons. For the medium-range errors, we even see an increase at the start of the experiments before they eventually decrease, after several dozens of cycles. At the end of the experiments, the forecast is significantly more accurate with \labelNN{} than with \labelSCNNa{}, which is what we hoped for. In some cases (first-guess and medium range), the forecast is even better with \labelNN{} than \labelSCNNt{} (NN trained offline with the truth). This results may seem at first somewhat surprising because, unless there has been some optimisation issues, the NN trained offline with the truth should provide the most accurate model error predictions. However, one must keep in mind that two essential simplifications have been made:
\begin{enumerate}
    \item the model error growth is linear in time (\cref{ssec:offline-nn-training});
    \item the model error correction is constant over the DA window (\cref{ssec:methodology-nn4dvar-simplified}).
\end{enumerate}
This explains why the NN trained offline with the truth is suboptimal in the DA and forecast experiments considered here. The first assumption could be circumvented by using samples of the true model error for a $\delta t=\SI{20}{\min}$ forecast (obviously, this would not be possible when training with the truth) but the second assumption is intrinsic to the simplified NN 4D-Var formulation. This second assumption allows us to build NN 4D-Var as a relatively simple extension of the currently implemented weak-constraint 4D-Var, but it has a negative impact on the forecast that we will illustrate in the following section.

\subsection{Focus on the first day of forecast}
\label{ssec:online-forecast-error-first-guess}

\begin{figure}[tbp]
    \centering
    \includegraphics{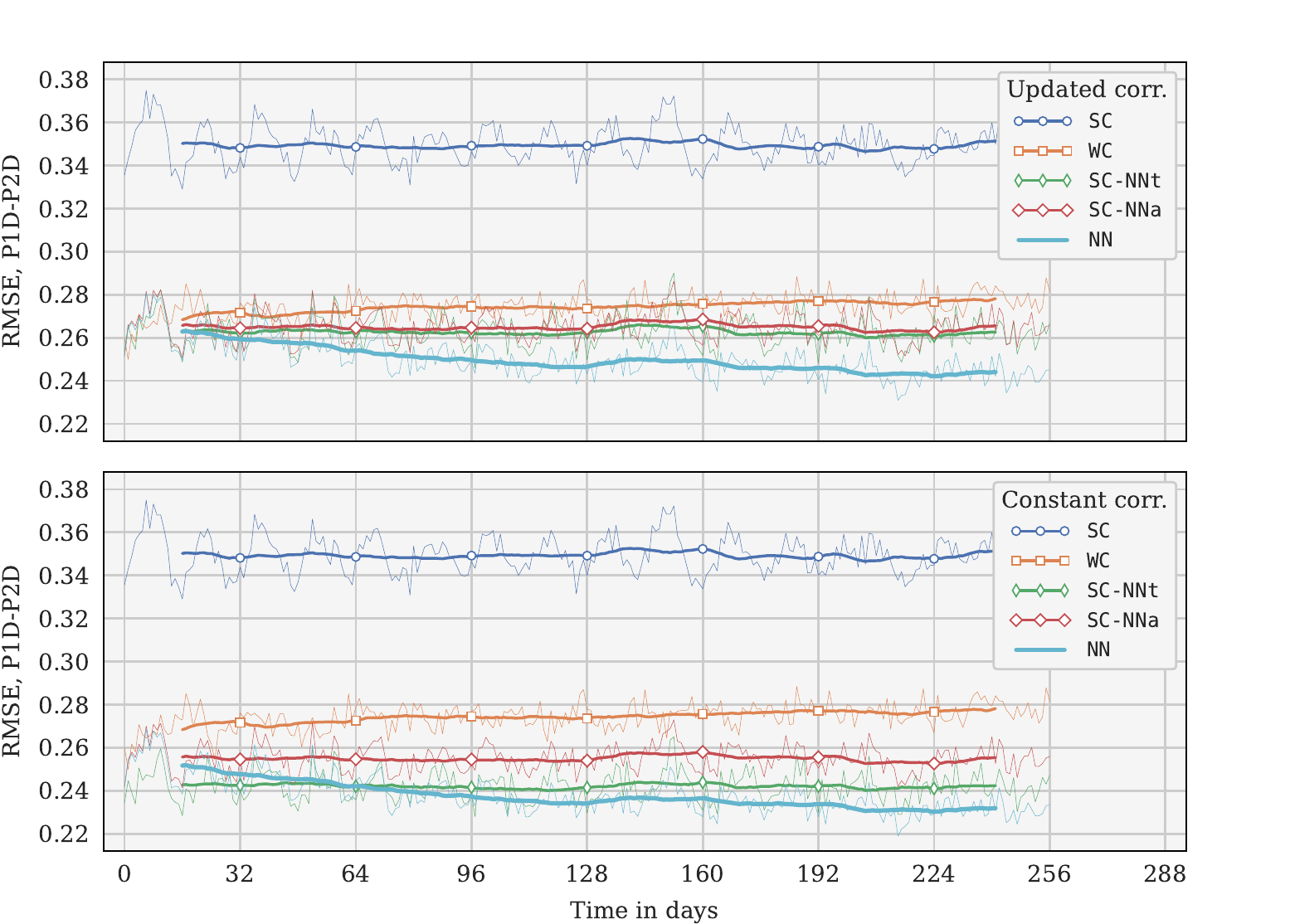}
    \caption{\label{fig:online-rmse-f} Forecast scores for the online experiments. Evolution of the forecast RMSE, averaged over the $\num{128}$ experiments and over P1D-P2D, as a function of time for the five 4D-Var variants: \labelSC{} in blue, \labelWC{} in orange, \labelSCNNt{} in green, \labelSCNNa{} in red, and \labelNN{} in teal. The NN correction is either updated every day (top panel, same as the middle panel of \cref{fig:online-rmse-nn}) or kept constant throughout the entire forecast (bottom panel). The thin lines report the instantaneous values and the thick lines report the running-average over $\num{32}$ cycles.}
\end{figure}

\Cref{fig:online-rmse-f} shows the temporal evolution of the errors in the second day of forecast in two cases: (i) the NN correction is updated every day (as has been done previously -- this corresponds to the first-guess errors) or (ii) it is kept constant throughout the entire forecast. The forecast errors with the NN (\labelSCNNt{}/\labelSCNNa{}/\labelNN{}) are systematically lower in the second case than in the first. Indeed, in the 4D-Var variants considered here, the NN correction is constant over the DA window, hence the forecast model is more consistent with the 4D-Var analysis when the NN correction is not updated. Of course, there is a limit to this logic because the model error evolves over time -- see the discussion on the accuracy of the forecast with \labelWC{} in \cref{ssec:offline-corrected-forecast} -- which is why it is important to update the NN correction for the forecast accuracy. Therefore, we believe that implementing NN 4D-Var without the assumption of a constant model error over the window should have a positive impact on the analysis, but also in the forecast. Rge implementation of such a formulation would not be trivial, as it could not be built directly on top of the existing WC 4D-Var. Although we have not attempted it in this study, we envisage considering it in further studies.

\subsection{Forecast errors at the end of the experiments}

\begin{figure}[tbp]
    \centering
    \includegraphics{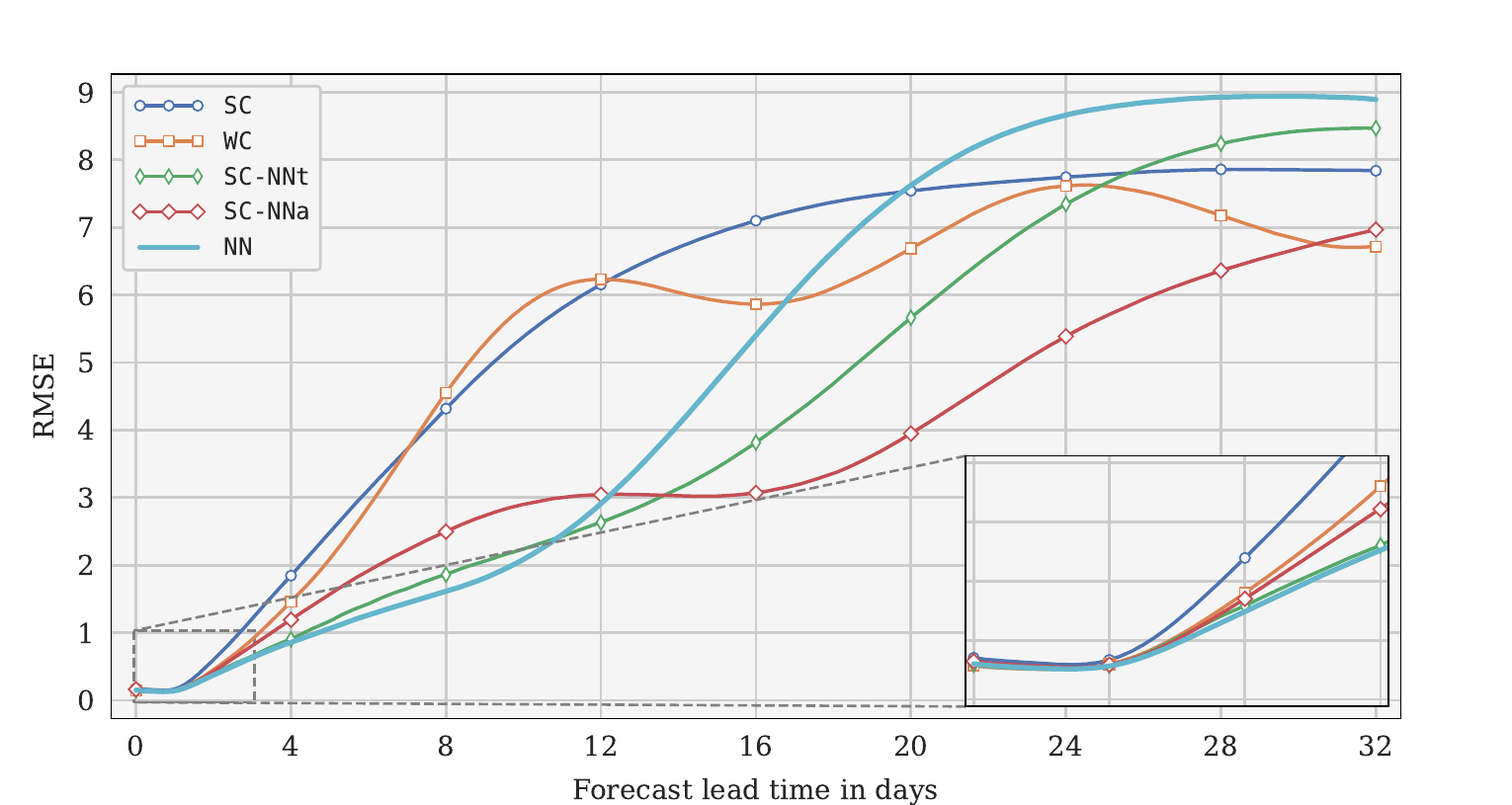}
    \caption{\label{fig:online-forecast}Online forecast results. Evolution of the time-averaged forecast RMSE, averaged over the last $\num{32}$ cycles and over the $\num{128}$ experiments, as a function of the forecast horizon for the five 4D-Var variants: \labelSC{} in blue, \labelWC{} in orange, \labelSCNNt{} in green, \labelSCNNa{} in red, and \labelNN{} in teal.The insert zooms in the short forecast lead times.}
\end{figure}

Finally, \cref{fig:online-forecast} shows the evolution of the forecast RMSE, averaged over the last $\num{32}$ cycles and over the $\num{128}$ repetitions of the experiments, as a function of the forecast lead time. The errors are the same as the ones in \cref{ssec:online-forecast-error-temporal-evolution,ssec:online-forecast-error-first-guess}, but aggregated and shown in a different way.
For the \labelNN{} variant, the forecast errors up to day $\num{10}$ are consistent with the description in \cref{ssec:online-forecast-error-temporal-evolution}. After day $\num{10}$, the forecast errors increase accelerate, which indicates that the NN correction is not any more valid. This is the same phenomenon as what has been described in \cref{ssec:offline-corrected-forecast} for \labelSCNNt{} and \labelSCNNa{}, but this time, the error increase is earlier and quicker. Once again, we believe that this result is related to the limited predictive power of the chosen NN. However, using a larger and deeper NN (\ie{} with more parameters) is not necessarily a good strategy with online learning. Indeed, based on preliminary experiments, we conclude that if the number of parameters is large, the background error covariance matrix for parameters (called $\mathbf{P}$ in \cref{ssec:methodology-nn4dvar}) must be small to avoid a quick divergence of the method. The downside of this choice is that it naturally slows down the learning process. This is why, with online learning, it is important to keep the number of parameters as small as possible, as explained by \citep{farchi-2021b}. Hence, the use of online learning could initially be limited to the correction of short-term forecasts.

\section{Conclusions}
\label{sec:conclusions}

In this article, we have developed a new variant of weak-constraint 4D-Var, in which a set of parameters can be jointly estimated alongside the system state. The new method is called NN 4D-Var to emphasise the fact that it is used in this article to estimate the coefficients (weights and biases) of a NN. It can be seen as a simplified variant of the original NN 4D-Var method introduced by \citet{farchi-2021b}, dedicated to model error correction. It is assumed that the NN provides a correction to a physical model, added after each integration, and constant over the DA window. These simplifications make the method very similar to the forcing formulation of weak-constraint 4D-Var, and hence easier to implement on top of an existing implementation of weak-constraint 4D-Var, such as the one available in the OOPS framework.

In the second part of the article, we have provided a numerical illustration of the new, simplified NN 4D-Var algorithm in conditions which are as close as possible to operational. The illustrations use twin experiments with OOPS-QG, a two-layer two-dimensional QG model. A simple yet non-trivial model error setup is introduced, where the layer depths and integration step of the model are perturbed. The model error correction is computed using a small, dense NN acting on vertical columns, like the one used for an operational model by \citet{bonavita-2020}. The NN is first trained offline, using the analyses and analysis increments of a DA experiment with the non-corrected model, following the method originally introduced by \citet{brajard-2020}. The corrected model is then used in forecast and DA experiments, and provides in both cases significant improvements in the scores as already shown by \citet{farchi-2021}. Then, the NN is trained online using the new, simplified NN 4D-Var algorithm. The results confirm the findings of \citet{farchi-2021b} for the original NN 4D-Var algorithm. With proper tuning of the background error covariance matrices, an online, joint estimation of the system state and the NN parameters is possible. As new observations become available, the model error correction becomes more accurate, which translates into lower analysis, first-guess, and short- to mid-term forecast errors than in the offline training case. 

The results also illustrate two limitations of the simplified NN 4D-Var method. The first is related to the assumption of a constant model error throughout the window. This is necessary to build the new method on top of an existing weak-constraint 4D-Var implementation, but we believe that relaxing this simplification could improve the analysis and short-term forecast errors. This could be the topic of further studies on the the method. The other limitation is somewhat more fundamental: the online training process is slower as the number of parameters to estimate is larger, as already highlighted by \citet{farchi-2021b}. This underlines the importance of choosing smart, parameter-efficient NNs.

At this point, we estimate that the simplified NN 4D-Var method is mature enough for more realistic applications, for example with the IFS. Implementing the new formulation in this operational model will only require developing an interface to the NN library with all the algorithmic developments already in place in the OOPS framework. For such application, we would typically use the vertical NN architecture of \citet{bonavita-2020}, for which the number of parameters is much lower than the number of system state variables. In this case however, the main difficulty would come from the fact that the true state of the system is unknown, which makes the evaluation much harder because the diagnostics should be based on observations. Nevertheless, we should be able to rely on the test suite developed by ECMWF to evaluate the potential benefits of proposed upgrades to the operational assimilation and forecast systems.

Finally, the current implementation of the simplified NN 4D-Var method in OOPS is dedicated to model error correction only, \ie{} the NN is trained for model error correction only. Nevertheless, there is no obstacle to use this method to train the NN for other tasks (\eg{} observation bias correction) provided that we are able to model their effect on the 4D-Var cost function.

\section*{Acknowledgements}

A. Farchi has benefited from a visiting grant of the ECMWF. CEREA is a member of Institut Pierre--Simon Laplace.

\bibliography{bibtex}

\end{document}